# On-Line Condition Monitoring using Computational Intelligence

C.B. VILAKAZI, T. MARWALA, P. MAUTLA and E. MOLOTO
School of Electrical and Information Engineering
University of the Witwatersrand
Private Bag 3, Wits, 2050
SOUTH AFRICA
t.marwala@ee.wits.ac.za

*Abstract:-* This paper presents bushing condition monitoring frameworks that use multi-layer perceptrons (MLP), radial basis functions (RBF) and support vector machines (SVM) classifiers. The first level of the framework determines if the bushing is faulty or not while the second level determines the type of fault. The diagnostic gases in the bushings are analyzed using the dissolve gas analysis. MLP gives superior performance in terms of accuracy and training time than SVM and RBF. In addition, an on-line bushing condition monitoring approach, which is able to adapt to newly acquired data are introduced. This approach is able to accommodate new classes that are introduced by incoming data and is implemented using an incremental learning algorithm that uses MLP. The testing results improved from 67.5% to 95.8% as new data were introduced and the testing results improved from 60% to 95.3% as new conditions were introduced. On average the confidence value of the framework on its decision was 0.92.

*Key-Words:* - Dissolve gas analysis (DGA), multi-layer perceptrons (MLP), radial basis function (RBF), support vector machines (SVM), Learn++

## 6   Introduction

Bushings are important components in transmission and distribution of electricity. The reliability of bushings affects the availability of electricity in an area as well as the economical operation of the area. Transformer failure studies show that bushings are among the top three most common causes of transformer failure [1][2]. Bushing failure is usually followed by a catastrophic event such as tank rupture, violent explosion of the bushing and fire [2]. In such eventuality the major concern is the risk of collateral damage and personnel injury.

Various diagnostic tools exist such as on-line partial discharge (PD) analysis, on-line power factor, and infrared scanning to detect an impending transformer failure [3]. However, few of these methods can in isolation provide all of the information that a transformer operator requires to decide upon a cause of action. Computational intelligence methods can be used in conjunction with the above-mentioned methods for bushing condition monitoring. Condition monitoring has a number of important benefits such as; an unexpected failure can be avoided through the possession of quality information relating to on-line condition of the plant and the consequent ability to identify faults while in incipient levels of development. In this paper, methods that are based on computational intelligence techniques are developed and then used for interpreting data from dissolve gas-in-oil analysis (DGA) test. The methods use machine learning classifiers multi-layer perceptrons (MLP), radial basis functions (RBF) and support vector machines (SVM). These methods are compared and the most effective method is implemented within the on-line framework. The justification for an on-line implementation is based on the fact that training data become available in small batches and that some new conditions only appear in subsequent data collection stage and therefore there is a need to update the classifier in an incremental fashion without compromising on the classification performance of the previous data.

## 7   Background

This section gives a background on dissolve gas analysis, artificial neural networks and support vector machines.

### 7.1   Dissolve gas analysis (DGA)

DGA is the most commonly used diagnostic technique for transformers and bushings [4][5]. DGA is used to detect oil breakdown, moisture presence and PD activity. Fault gases are produced by degradation of transformer and bushing oil and solid insulation such as paper and pressboard, which are all made of cellulose [6]. The gases produced from the transformer and bushing operation are [5][7][8]: (1) Hydrocarbons gases and hydrogen: methane ($CH_4$), ethane ($C_2H_6$), ethylene ($C_2H_4$), acetylene ($C_2H_2$) and hydrogen ($H_2$); (2) Carbon oxide: carbon monoxide (CO) and carbon dioxide ($CO_2$); and (3) Non-fault gases: nitrogen ($N_2$) and oxygen ($O_2$).

The causes of faults are classified into two main groups, which are partial discharges and thermal heating. Partial discharges faults are divided into high-energy discharge and low-energy discharge. High-energy discharge is known as arcing and low energy discharge is referred to as corona. The quantity and types of gases reflect the nature and extent of the stressed mechanism in the bushing [9]. Oil breakdown is shown by the presence of hydrogen, methane, ethane, ethylene and acetylene. High levels of hydrogen show that the degeneration is due to corona. High levels of acetylene occur in the presence of arcing at high temperature. Methane and ethane are produced from low temperature thermal heating of oil and high temperature thermal heating produces ethylene, hydrogen as well as a methane and ethane. Low temperature thermal degradation of cellulose produces $CO_2$ and high temperature produces CO.

## 7.2 Artificial neural network

Artificial neural networks (ANN) are data processing systems that learn complex input-output relationships from data [10]. A typical ANN consists of simple processing elements called neurons that are highly interconnected in an architecture that is inspired by the structure of biological neurons in human brain [10]. There are different types of ANN models; two that are commonly used and considered in this paper are multi-layer perceptrons (MLP) and radial basis functions (RBF) networks.

### 7.2.1 Multi-layer perceptrons

MLPs are feed-forward neural networks that provide a general framework for representing non-linear functional mappings between a set of input variable and a set of output variables. This is achieved by representing a non-linear function of many variables in terms of a composition of non-linear single variable, called activation functions [10]. Fig.1 shows the architecture of an MLP with four input layer neurons, three hidden layer neurons and two output layer neurons. From Fig.1, the relationship between input (x) and output (y) can be written as [10]:

$$y_j = \bar{f}\left(\sum_{j=0}^{N} w_{kj}^{2} f\left(\sum_{i=0}^{d} w_{ji}^{1} x_i\right)\right) \quad (1)$$

In (1), $w_{kj}^{(2)}$ and $w_{ji}^{(1)}$ represents the weight of the layer 2 and layer 1 respectively while $N$ and $d$ are the number of input layer neurons and output layer neurons, respectively.

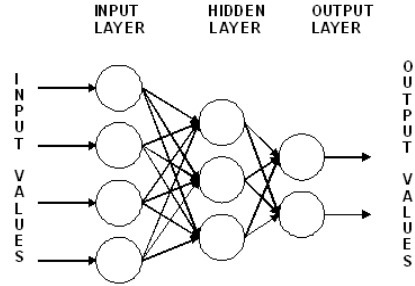

Fig.1. Architecture of an MLP [10]

The aim of training is to minimize the error function in order to find the most probable weight connection given the training data. MLP training teaches the network to match the input to a corresponding output. Two types of learning algorithms exist, supervised and unsupervised learning. Supervised learning is used in this paper to estimate the weight parameters. In supervised learning, the neural network is presented with both the input and output values. The actual output of MLP together with its associated target is used to evaluate the error function to quantify the error of the mapping [10]. The goal of parameter estimation is to minimize the prediction in equation 1 and the target $t$. This is achieved by minimising the cross-entropy error $(E)$ [10]:

$$E = -\beta \sum_{n=1}^{N}\sum_{k=1}^{K}\{t_{nk}\ln(y_{nk}) + (1-t_{nk})\ln(1-y_{nk})\} + \frac{\alpha}{2}\sum_{j=1}^{W} w_j^2 \quad (2)$$

In (2), the cross-entropy function is chosen because it has been found to be ideally suited to classification problems. In equation 2, $n$ is the index for the training pattern, $\beta$ is the data contribution to the error and $k$ is the index for the output units. The second term in equation 2 is the regularisation parameter and it penalises weights of large magnitudes. This regularisation parameter is called the weight decay and its coefficient, $\alpha$, determines the relative contribution of the regularisation term on the training error. This regularisation parameter ensures that the mapping function is smooth. Including the regularisation parameter has been found to give significant improvements in network generalisation [10]. In this paper to minimise equation 2, a method called scaled conjugate gradient method is used [11].

### 7.2.2 Radial basis function

RBFs are type feed-forward neural networks employing a hidden layer of radial units and an output layer of linear units [10]. In RBF, the distance between the input vector and output vector determines the activation function [10]. RBF have their roots in techniques of performing exact interpolation of a set of data points in a multi-dimensional space. This interpolation requires that every input target be mapped exactly onto corresponding target vector. Fig.2 shows the architecture of RBF with four input layer neurons, five hidden layer neurons and two output layer neurons.

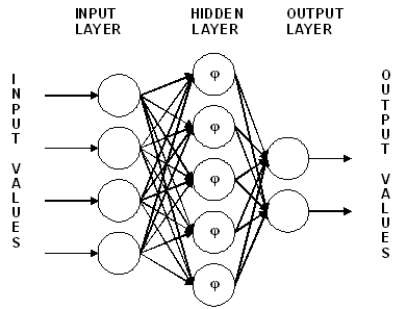

Fig.2. Architecture of RBF [10]

The RBF network can be mathematically described as follows [10]:

$$f_k(x) = \sum_{j=0}^{n} w_{kj} \phi_j(x) + b_k \qquad (3)$$

where $f_k$ represents the k-th output layer transfer function, $w$ and $b$ represents the weights and biases, and $\phi_j$ represents the j-th input layer transfer function represented in this paper by:

$$\phi_j(x) = \exp\left(-\frac{\|x - \mu_j\|^2}{2\sigma_j^2}\right) \qquad (4)$$

Here $x$ represents the input, $\mu$ represents the fixed centre position and $\sigma$ represents fixed variance. RBF consists of two stage training technique, in the first stage the input data are used to determine the parameters of the basis function. The basis functions are kept fixed while the second-layer weights are found in the second level of training. The second level training is explained in the next section.

In the first stage, the input data are used to determine the centre and variance of the basis functions. A randomly selected subset of the input training data set is selected for use as the basis centres. Clusters of training data are then identified and a basis function is centred at each cluster. The parameter $\mu$ is then chosen to be maximum distance between the basis function centres. In the second stage of training, the basis functions are kept fixed and the output layer weight is modified and this is equivalent to a single-layer neural network.

### 7.3 Support vector machines

SVM is a learning approach that implements the principle of Structural Risk Minimization (SRM). Structural risk minimization principle has been observed to be superior to the empirical risk minimization principle used in conventional neural networks [12]. SVM was developed to solve classification problems [12] and is schematically represented in Fig. 3. The idea behind SVM is to map an input space, x into a higher dimensional feature space, z. The goal is to find a kernel function ($f(x)$) that will map the input space to the training inputs to training outputs. Various feature spaces are used such as polynomial, Gaussian, Fourier series, splines as well as RBF and MLP nested within the activation function [13].

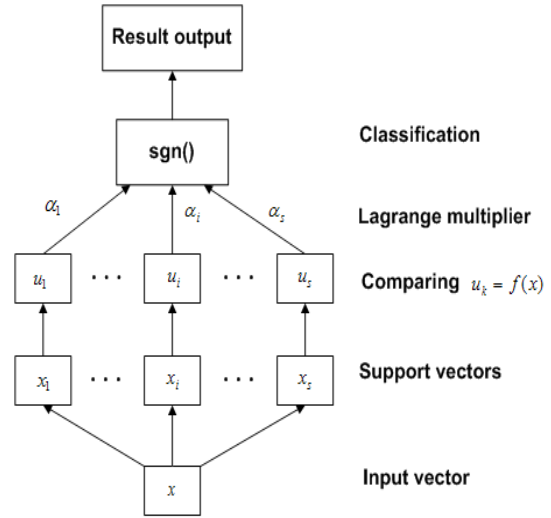

Fig.3. Mechanism of SVM

The classification inner activation function is given by [13]:

$$f(x) = \sum_{i=1}^{m} (\alpha - \alpha_i{}^*) \cdot k(x_i) \qquad (5)$$

Here, $k(x_i)$ is the kernel function and $\alpha_i$, $\alpha_j{}^*$ are the Langrage multipliers. The hyperplane that optimally separates the data are derived by minimizing the

Langrangian, $\Phi$ with respect to the weights $w$, bias $b$ and $\alpha$ [13] given by:

$$\phi = \frac{1}{2}\|w\|^2 - \sum \alpha_i \left(y_i\left[\langle w, x_i\rangle + b\right] - 1\right) \quad (6)$$

The multipliers are constrained by $0 \leq \alpha_i, \alpha_i^* \geq C$ where $C$ is the misclassification tolerance or capacity. If the value of $C$ is too large, the kernel function will over-fit the training data and will not have good generalization properties.

# 8 Proposed frameworks

The proposed frameworks for fault diagnosis are a two-level implementation. The first level of the diagnosis identifies if the bushing is faulty or not. If the bushing is faulty, the second level determines the types of faults, which are thermal fault, PD faults and faults caused by an unknown source. Generally, the procedure of fault diagnosis includes three steps, extracting feature and data pre-processing, training the classifiers and identifying transformer fault with the trained classifiers. Fig.4 shows the block diagram of the proposed methodology.

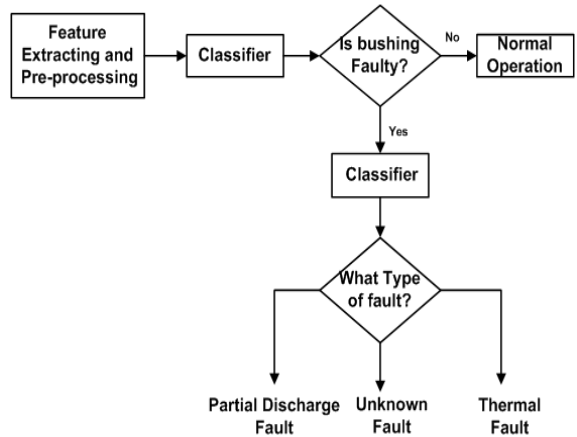

Fig.4. Block diagram of the proposed methodology

## 8.1 Data processing

DGA is used to determine the faulty gases in the bushing oil. The content information reflects the states of the transformer and bushing. Ten diagnostic gases mentioned in Section 2 are extracted, which are $CH_4$, $C_2H_6$, $C_2H_4$, $C_2H_2$, $H_2$, CO, $CO_2$, $N_2$, $O_2$ and total dissolved combustible gases. The total dissolved combustible gas is given by the sum of methane, hydrogen, acetylene, ethane, ethylene and hydrogen. The faulty gases are analysed using the IEEE C57.104 standards [14]. Data pre-processing is an integral part of neural network architecture. Data pre-processing makes it easier for the network to learn. Data are normalized to fall within 0 and 1, using linear normalization.

## 8.2 MLP, RBF and SVM classifiers

In this method MLP, RBF and SVM are trained. As the classifier for the first level and second level classifications are trained in a similar way, the training procedure explained below applies to both the two-class and three class problems.

The optimal number of hidden layer was found by using exhaustive search. An MLP with optimal number of hidden layer neuron was trained using the scaled conjugate gradient [11]. The centroids of RBF are found by using the Gaussian mixture model with circular covariance using the Expectation Maximization (EM) algorithm [10] and the output layer was trained using the scaled conjugate gradient algorithm. Cross-validation is used to ensure that a network with good generalization property is achieved. Cross-validation was used to determine the best kernel function and capacity of the SVM. Then SVM was trained with the optimal hyperplane and the optimal C.

The networks are tested with 1000 data points that are randomly selected from the data set. Table 1 shows the results of the networks trained to identify if the transformer is faulty or not which is called second level stage. The methods are tested using specificity and sensitivity. Sensitivity is defined as the probability of the classifier predicting faults correctly and specificity is the probability of a classifier predicting the non-faulty state correctly.

Table 1: Comparison of the performance of different frameworks for first level of fault diagnosis

|  | MLP | RBF | SVM |
|---|---|---|---|
| **Accuracy (%)** | 98.9 | 97.4 | 98.5 |
| **Specificity** | 0.796 | 1.000 | 0.996 |
| **Sensitivity** | 0.999 | 0.885 | 0.885 |
| **Training Time(s)** | 41.236 | 0.625 | 1975.437 |
| **Classification Time(s)** | 0.0157 | 0.0314 | 104.314 |

The table compares the framework in terms of accuracy, training and testing time. MLP classifier shows classification accuracy of 98.9%, RBF shows 97.4% and SVM gives 98.5% classification accuracy. This table shows that there is no significant difference between SVM and MLP classifiers. Although, RBF performs worse than MLP and SVM in terms of classification accuracy, it trains faster while SVM is computationally most expensive.

Table 2 compares the results of the networks designed in terms of accuracy, training time and testing time to classify bushing conditions into thermal fault, PD faults and faults caused by an unknown source bushing faults and this is called second level classification. This table shows that the MLP classifier gives 98.62% classification accuracy while RBF and SVM classifier give 81.73% and 96.9%, respectively. In the second level classification, the MLP classifier performs better than the RBF and SVM.

Table 2: Comparison of the performance of different frameworks for second level of fault diagnosis

|  | MLP | RBF | SVM |
|---|---|---|---|
| **Accuracy (%)** | 98.62 | 81.73 | 96.90 |
| **Training Time (s)** | 30.016 | 1.038 | 83.906 |
| **Classification Time (s)** | 0.780 | 0.05 | 1.094 |

Because the MLP gives the best results compared with RBF and MLP on condition monitoring of bushings, it is therefore selected as the most appropriate learning engine and thus it is chosen for on-line learning.

## 9 On-line learning

As indicated earlier in the paper, on-line learning is suitable for modelling dynamically time-varying systems, where the operating region changes with time. It is also suitable, if the data available is not adequate and does not fully represent the system. Another advantage of on-line learning is that it is able to accommodate new conditions that may be introduced by incoming data. An on-line bushing condition monitoring system must have incremental learning capability if it is to be used for automatic and continuous on-line monitoring. The on-line bushing monitoring system improves reliability, reduces maintenance cost and minimizes out-of-service time for a transformer. The basis of on-line learning is incremental learning, which has been studied by a number of researchers [15][16][17][18]. The difficulty in on-line learning is the tendency of an on-line learner to forget information gathered during the early stages of learning [19]. The method of on-line learning adopted in this paper is Learn++ [20].

### 4.1. Learn++

Learn++ is an incremental learning algorithm that uses an ensemble of classifiers that are combined using weighted majority voting [19]. Learn++ was developed by Polikar [19] and was inspired by a boosting algorithm called adaptive boosting (AdaBoost). Each classifier is trained using a training subset that is drawn according to a distribution. The classifiers are trained using a weakLearn algorithm. The requirement for the weakLearn algorithm is that it must be able to give a classification rate of less than 50% initially [20]. For each database $D_k$ that contains training sequence, $S$, where $S$ contains learning examples and their corresponding classes, Learn++ starts by initialising the weights, $w$, according to the distribution $D_T$, where $T$ is the number of hypothesis. Initially the weights are initialised to be uniform, which gives equal probability for all instances to be selected to the first training subset and the distribution is given by:

$$D = 1/m \qquad (7)$$

where $m$ represents the number of training examples in $S_k$. The training data are then divided into training subset $T_R$ and testing subset $T_E$ to ensure weakLearn capability. The distribution is then used to select the training subset $T_R$ and testing subset $T_E$ from $S_k$. After the training and testing subset have been selected, the weakLearn algorithm is implemented. The weakLearner is trained using subset, $T_R$. A hypothesis, $h_t$, obtained from weakLearner is tested using both the training and testing subsets to obtain an error, $\varepsilon_t$:

$$\varepsilon_t = \sum_{t:h_t(x_i)\neq y_i} D_t(i) \qquad (8)$$

The error is required to be less than 0.5; a normalized error $\beta_t$ is computed using:

$$\beta_t = \varepsilon_t / 1 - \varepsilon_t \qquad (9)$$

If the error is greater than 0.5, the hypothesis is discarded and new training and testing subsets are selected according to $D_T$ and another hypothesis is computed. All classifiers generated are then combined using weighted majority voting to obtain composite hypothesis, $H_t$:

$$H_t = \arg\max_{y\in Y} \sum_{t:h_t(x)=y} \log(1/\beta_t) \qquad (10)$$

Weighted majority voting gives higher voting weights to a hypothesis that performs well on the training and testing subsets. The error of the composite hypothesis is computed by:

$$E_t = \sum_{t:H_i(x_i) \neq y_i} D_t(i) \quad (11)$$

If the error is greater than 0.5, the current hypothesis is discarded and the new training and testing data are selected according to the distribution $D_T$. Otherwise, if the error is less than 0.5, the normalized error of the composite hypothesis is computed as:

$$B_t = E_t / 1 - E_t \quad (12)$$

The error is used in the distribution update rule, where the weights of the correctly classified instances are reduced, consequently increasing the weights of the misclassified instances. This ensures that instances that were misclassified by the current hypothesis have a higher probability of being selected for the subsequent training set. The distribution update rule is given by

$$w_{t+1} = w_t(i) \times B_t^{1-[|H_t(x_i) \neq y_i|]} \quad (13)$$

Once the $T$ hypothesis is created for each database, the final hypothesis is computed by combining the hypothesis using weighted majority voting given by:

$$H_t = \arg\max_{y \in Y} \sum_{k=1}^{K} \sum_{t:H_t(x)=y} \log(1/\beta_t) \quad (14)$$

### 4.2. Confidence measurement

A simple procedure is used to determine the confidence of the algorithm on its own decision. A vast majority of hypothesis agreeing on a given instances can be interpreted as an algorithm having confidence on the decision. Let us assume that a total of $T$ hypothesis are generated in $k$ training sessions for a C-class problem. For any given example, the final classification class, if the total vote class c receives is given by [21][22]:

$$\xi_c = \sum_{t:h_t(x)=c} \psi_t \quad (15)$$

where $\psi_t$ denotes the voting weights of the $t^{th}$ hypothesis $h_t$. Normalizing the votes received by each class gives:

$$\gamma_c = \xi_c / \sum_{c=1}^{C} \xi_c \quad (16)$$

$\gamma_c$ can be interpreted as a measure of confidence on a scale of 0 to 1. A high value of $\gamma_c$ shows high confidence in the decision and consequently, a low value of $\gamma_c$ shows low confidence in the decision. It should be noted that the $\gamma_c$ value does not represent the accuracy of the results but the confidence of the system on its own decision.

## 10 Experimental results for on-line learning

The first experiment evaluates the incremental capability of the algorithm using first level fault diagnosis. The data used were collected from bushing over a period of 2.5 years from bushings in services. The algorithm is implemented with 1500 training examples and 4000 validation examples. The training data were divided into five databases each with 300 training instances. In each training session, Learn++ is provided with each database and 20 hypotheses are generated. The weakLearner uses an MLP with 10 input layer neurons, 5 hidden layer neurons and one output layer neuron. To ensure that the method retains previously learned data, the previous database is tested at each training session. The first row of Table 3 shows the performance of the Learn++ on the training data for different databases. On average, the weakLearner gives 60% classification rate on its training dataset, which improves to 98% when the hypotheses are combined. This demonstrates the performance improvement of Learn++ as inherited from AdaBoost on a single database. Fig. 5 shows the performance of Learn++ on training dataset against the number of classifiers for a single database. Each column shows the performance of current and previous databases and this is to show that Learn ++ does not forget previously learned information, when new data are introduced. The last row of Table 3 shows the classifiers performances on the testing dataset, which gradually improves from 65.7% to 95.8% as new databases become available, demonstrating incremental capability of Learn++. Fig. 6 shows the performance of Learn++ on one dataset against the number of datasets. Table 4 shows that the confidence of the framework increases as new data are introduced.

The second experiment was performed to evaluate whether the frameworks can accommodate new classes. The faulty data were divided into 1000 training examples and 2000 validation data, which contained all the three classes. The training data were divided into five databases, each with 200 training instances. The first and second database contained, training examples of PD and thermal faults.

The data of unknown fault were introduced in training session three. In each training session, Learn++ was provided with each database and 20 hypotheses were generated. The last row of Table 3 shows that the classifiers performances increase from 60% to 95.3% as new classes were introduced in the subsequent training datasets. Table 5 shows the training and testing performance of the algorithm as new conditions are introduced.

Table 3: Performance of Learn++ for first level on-line condition monitoring, key: S =databases.

| Dataset | S1 | S2 | S3 | S4 | S5 |
|---|---|---|---|---|---|
| S1 | **89.5** | 85.8 | 83.00 | 86.9 | 85.3 |
| S2 | — | **91.4** | 94.2 | 93.7 | 92.9 |
| S3 | — | — | **93.2** | 90.1 | 91.4 |
| S4 | — | — | — | **92.2** | 94.5 |
| S5 | — | — | — | — | **98.0** |
| Learn++ Testing (%) | 65.7 | 79.0 | 85.0 | 93.5 | 95.8 |

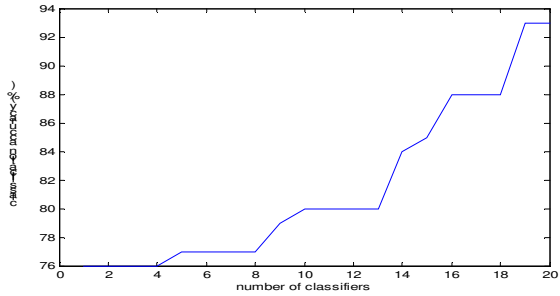

Fig.5. Performance of Learn++ on training data against the number of classifiers

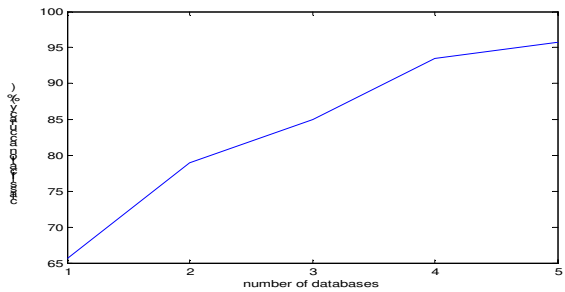

Fig.6. Performance of Learn++ on testing data against the number of databases

The final experiment addressed the problem of bushing condition monitoring using MLP network trained using batch learning. This was done to compare the classification rate of Learn++ with that of an MLP.

Table 4: Confidence values of the algorithm for classified cases for all databases and classes.

|  | S1 | S2 | S3 | S4 | S5 |
|---|---|---|---|---|---|
| **Classified:** |  |  |  |  |  |
| **Normal Class** | 0.66 | 0.85 | 0.92 | 0.94 | 0.94 |
| **Faulty Class** | 0.49 | 0.64 | 0.81 | 0.90 | 0.90 |

Table 5: Performance of Learn++ for second stage bushing condition monitoring. key: S + databases

| Dataset | S1 | S2 | S3 | S4 | S5 |
|---|---|---|---|---|---|
| S1 | **95.0** | 95.2 | 94.6 | 95.7 | 95.1 |
| S2 | — | **96.3** | 96.0 | 96.8 | 95.3 |
| S3 | — | — | **97.0** | 96.4 | 96.5 |
| S4 | — | — | — | **97.8** | 96.8 |
| S5 | — | — | — | — | **99.2** |
| Testing (%) | 60.0 | 65.2 | 76.0 | 83.0 | 95.3 |

An MLP with the same set of training example as Learn++ was trained and the trained MLP was tested with the same validation data as Learn++. This was done for the first and second levels fault classification. The first level fault diagnosis, the MLP gave classification rate of 97.2% whereas the second level MLP give a classification rate of 96.0%. This is when the classifier had seen all the fault classes in priori. If the classifier had not seen all the fault cases, the performance decreases from 65.7% for database 1 to 30.0% for database 2 to 3 for the first level fault classification. However, for second level fault classification it decreases from 60.0% to 22.3%.

## 6   Discussion and conclusions

The data from DGA are interpreted using machine, learning classifiers MLP, RBF and SVM to detect impending bushing failure. The first level of fault diagnosis evaluates if the transformer is faulty or not while the second level determines the nature of the faults. It is observed that there is no major difference in the performance of the MLP and SVM classifiers for the first level of fault diagnosis. However, SVM takes longer to train than the MLP. Although the RBF is faster to train, it gives the worst results of the three classifiers. As the MLP gives the best performance of the three classifiers, it is used to implement on-line learning. An on-line learning method that uses incremental learning is implemented for on-line bushing condition monitoring. Experimental results of the on-line learning method demonstrate that the proposed framework has incremental learning capability.

Furthermore, the results show that the framework is able to accommodate new conditions introduced by incoming data. The results further show that the algorithm has a high confidence in its own decision.